\documentclass[11pt]{article}
\usepackage[utf8]{inputenc}
\usepackage[T1]{fontenc}
\usepackage[english]{babel}
\usepackage[margin=2.8cm]{geometry}
\usepackage[scaled=.95]{helvet}
\usepackage[numbers]{natbib}

\usepackage[dvips]{graphicx}
\usepackage{amsmath,amssymb,color}
\usepackage[utf8]{inputenc} 
\usepackage{hyperref}
\input epsf

\def\build#1_#2^#3{\mathrel{
\mathop{\kern 0pt#1}\limits_{#2}^{#3}}}

\def\Annexe{{\large Annexe}}

\newtheorem{rmq}{Remark}[section]

\newcommand{\bqa}{\begin{eqnarray}}
\newcommand{\eqa}{\end{eqnarray}}
\newcommand{\bdesc}{\begin{description}}
\newcommand{\edesc}{\end{description}}
\newcommand{\bqan}{\begin{eqnarray*}}
\newcommand{\eqan}{\end{eqnarray*}}

\def\norm#1{\left\| #1 \right\|}

\def\pa#1{\left( #1 \right)}

\def\cro#1{\left[ #1 \right]}
\def\acco#1{\left \{  #1\right \}}

\def\ndx2{n\,d_x^2}

\def\bkV{{\mathbb{V}}}

\def\bkF{\mbox{{\rm I\kern-.17em F}}}

\def\bkNp{\mbox{{\small\rm I\kern-.20em N}}}
\def\bkO{{\rm \kern.24em
            \vrule width.05em height1.4ex depth-.05ex
            \kern-.26em O}}
\def\bkZ{\mbox{{\rm Z\kern-.32em Z}}}

\def\non{\mbox{$^{\mbox{\bf --}}$\kern-.1em \vrule width.05em height1.5ex}}

\def\Sigmabar{\overline{\Sigma}}

\def\Var{\mbox{\rm $\bkV$ar}}

\def\emptysq{\mathbin{\vbox{\hrule\hbox{\vrule height1ex \kern.5em 
       \vrule height1ex}\hrule}}}

\usepackage{indentfirst}
\usepackage{bm}
\usepackage{multirow}
\usepackage{float}

\newtheorem{theoreme}{Theorem}

\newtheorem{lemma}{Lemma}
\usepackage{dsfont}
\usepackage{subfigure}
\usepackage{enumerate}

\begin{document}
\title{Online stochastic Newton methods for estimating the geometric median and applications}
\author{Antoine Godichon-Baggioni$^{(*)}$ and 
Wei Lu$^{(**)}$ \\
       $(*)$ Sorbonne Université, Laboratoire de Probabilités, Statistique et Modélisation  \\ 
       $(**)$ INSA Rouen Normandie, Laboratoire de Mathématiques de l'INSA \\     
         antoine.godichon$\_$baggioni@upmc.fr,  wei.lu@insa-rouen.fr \\
         \date{}
}
\maketitle
\begin{abstract}
In the context of large samples, a small number of individuals might spoil basic statistical indicators like the mean. It is difficult to detect automatically these atypical individuals, and an alternative strategy is using robust approaches. This paper focuses on estimating the geometric median of a random variable, which is a robust indicator of central tendency. In order to deal with large samples of data arriving sequentially, online stochastic Newton algorithms for estimating the geometric median are introduced and we give their rates of convergence. Since estimates of the median and those of the Hessian matrix can be recursively updated, we also determine confidences intervals of the median in any designated direction and perform online statistical tests.
\end{abstract}
\section{Introduction}

Large samples of observations are now commonplace due to advancements in measurement technology and improved computer storage capabilities. In such a large sample context, even a small number of individuals might spoil basic statistical indicators like the mean. Detecting automatically these atypical individuals is difficult, and adopting robust approaches is an appealing alternative. It is well known that the median is a robust indicator of central tendency, and here we concentrate on the geometric median of a random variable in $\mathbb{R}^p$. The geometric median, also called spatial median or multivariate $L_1$ median, is firstly introduced in \cite{gini1929di} and \cite{haldane1948note}. It is defined as the minimizer of $L_1$ distances to observations of a random variable. It has nice robustness properties such as a breakdown point at $0.5$ \cite{lopuhaa1991breakdown,kemperman1987median}.

Recently, the geometric median attached more attention in the field of machine learning. For example, in \cite{lu2017robust}, authors proposed a $L_1$ median filter as a tool of their mesh denoising method, which helps their model to preserve geometric features; an optimization algorithm of section line extraction was established based on geometric median \cite{zhang2021new}, the author considered the geometric median because it has the characteristics of noise immunity; an image filtering algorithm \cite{church2008spatial} has been proposed based on a spatial median filter, which shows better performances than mean filters, since the median is more robust than the mean when noisy pixels are present in the image.

In this paper, we focus on the estimation of the geometric median. An iterative algorithm called Weiszfeld's algorithm has been developed  \cite{weiszfeld1937point,gower1974algorithm,vardi2000multivariate}, and the method has been improved in \cite{beck2015weiszfeld}. The algorithm is simple and fast, but the procedure is not adapted in the case where data are acquired sequentially from files too large to be loaded into memory. To overcome this, and since the geometric median is defined as the minimizer of a convex function, an averaged stochastic gradient algorithm for estimating the geometric median has been proposed in \cite{cardot2013efficient}. However, as a first-order algorithm, in practice it can be very sensitive to the Hessian structure of the function to minimize \cite{bercu2020efficient,boyer2022asymptotic}.

In order to overcome this, we propose here new stochastic Newton type algorithms for estimating the geometric median. One difficulty encountered by stochastic Newton algorithms is the update of the inverse of the Hessian estimates. Our recursive estimation of the inverse of the Hessian is based on the Sherman-Morrison formula \cite{duflo1997random}, which avoids an expensive inverse matrix calculation. In order to overcome possible initialization problems, we also propose a weighted averaged version \cite{boyer2022asymptotic}. Thanks to the asymptotic efficiency of the algorithms, and since one can recursively estimate the covariance matrix, we introduce online confidence intervals of the geometric median in a chosen direction and to perform online statistical hypothesis tests.

The paper is organized as follows: we describe the general framework and explain the method for estimating recursively the inverse  of the Hessian in Section \ref{sec::framework}. In Section \ref{sec::algos} we present stochastic Newton algorithms and we state their rates of convergence. A simulation study for comparing the performances of different algorithms is also  given. Section \ref{ICandTests} is devoted to establishing recursive confidence intervals and performing online statistical tests for the geometric median. The proofs are gathered in Section \ref{proof}.

\section{Framework} \label{sec::framework}
\subsection{General framework}
The geometric median $m$ of a random variable $X$ taking values in $\mathbb{R}^p$ is the minimizer of the convex function $G:\mathbb{R}^p \longrightarrow \mathbb{R}$ defined for all $h \in \mathbb{R}^p $ by \cite{haldane1948note}
$$ G(h) =:\mathbb{E}\cro{g(X,h)}=\mathbb{E}\cro{\norm{X-h}-\norm{X}}.$$
Note that this definition does not assume the uniqueness of the median or the existence of the first order moment of $\norm{X}$.
From now on we suppose that following assumptions are fulfilled. \\
\begin{itemize}
\item \textbf{Assumption 1.} 
 The random variable $X$ is not concentrated around single points : there exists $C_6 > 0$ such that for all $ h \in \mathbb{R}^p$,
 \[
 \mathbb{E}\cro{\frac{1}{\norm{X-h}^6}} \le C_6. 
 \]
\item \textbf{Assumption 2.}
The random variable $X$ is not concentrated on a straight line : for all $ h \in \mathbb{R}^p$, there exists $h' \in \mathbb{R}^p$ such that $\langle h,h'\rangle \neq 0$ and 
\[
\Var\cro{\langle X,h'\rangle} > 0.
\]
\end{itemize}
Note that in Assumption 1, the order is usually obtained to be 2 in the literature \cite{cardot2013efficient}, and we increase the order to 6 for technical reasons, i.e. it is used to obtain the convergence rate of the Hessian's estimates that will be presented later. According to \cite{kemperman1987median},  Assumption 2 ensures that the function $G$ is strictly convex, so that the median $m$ is uniquely defined.
As shown in \cite{cardot2013efficient}, the function $G$ is differentiable everywhere, and one can check that the gradient is defined for all $h\in \mathbb{R}^p $ by:
$$\nabla G(h) = - \mathbb{E}\cro{\frac{X-h}{\norm{X-h}}}.$$
Moreover, the function $G$ is twice differentiable everywhere and its Hessian is given by \cite{koltchinskii1997m}
$$\nabla^2 G(h) = \mathbb{E}\cro{\frac{1}{\norm{X-h}}\pa{I_p - \frac{(X-h)(X-h)^T}{\norm{X-h}^2}}}.$$
 According to \cite{cardot2013efficient}, $\nabla^2 G(h)$ is positive definite under Assumptions 1 and 2. This is of particular interest to stochastic Newton type algorithms for estimating the geometric median, in which the information given by the Hessian matrix of the function $G$ will be taken into account.

\subsection{Some recalls on the averaged stochastic gradient algorithm}
An averaged stochastic gradient algorithm has been proposed in \cite{cardot2013efficient} for estimating the geometric median.
Given $X_1$, $X_2,  \ldots , X_n , X_{n+1}, \ldots, $, i.i.d copies of $X$, the stochastic gradient algorithm is given by
\begin{equation}\label{defASGD1}
m_{n+1}^{(SG)} = m_n^{(SG)} + \gamma_n\frac{X_{n+1}-m_n^{(SG)}}{\norm{X_{n+1}-m_n^{(SG)}}},  
\end{equation}
where $\gamma_n$ is a sequence of descent steps. Its averaged version consists of averaging all the estimated past values, which is defined recursively by 
\begin{equation}\label{defASGD2}
\overline{m}_{n+1} = \overline{m}_{n} + \frac{1}{n+1}\pa{m_{n+2}^{(SG)}-\overline{m}_{n}}
\end{equation}
with $m_{0}^{(SG)}$ bounded and $\overline{m}_{0} = m_{0}^{(SG)}$. Thus, the estimation can be easily updated. This algorithm has been deeply studied: its asymptotic efficiency is given in \cite{cardot2013efficient}, while the $L^p$ rates are derived in \cite{godichon2016estimating}. Moreover, the non-asymptotic behavior of this algorithm has also been studied in \cite{cardot2017online} by giving non-asymptotic confidence balls based on the derivation of improved $L^2$ rates of convergence. However, it's a first-order algorithm and thus can be very sensitive to the structure of Hessian of the function we try to minimize \cite{boyer2022asymptotic}, which means here that the random variable $X$ is fairly concentrated towards a straight line for instance.
\subsection{How to estimate the inverse of the Hessian}\label{How_invH}
We will then focus on Newton type methods, which is more adapted to deal with ill-conditioned problems. A major difficulty encountered by stochastic Newton algorithm is that we must be able to update the inverse of the Hessian estimate with a cost, in terms of computation time, as low as possible. We explain now how to estimate the inverse of the Hessian $\nabla^{2}G(m)$ in a recursive way when $m$ is known. The idea is to obtain an estimate of the form $\frac{1}{n}\sum_{k=1}^na_k\phi_k\phi_k^T$ to apply Riccati's formula \cite{duflo1997random}. We have
$$\nabla^2 g(X,h) = \frac{1}{\norm{X-h}}\pa{I_p - \frac{(X-h)(X-h)^T}{\norm{X-h}^2}}.$$ 
Note that $\pa{I_p-\frac{(X-h)(X-h)^T}{\norm{X-h}^2}}^2 = I_p-\frac{(X-h)(X-h)^T}{\norm{X-h}^2}$, we then have \begin{equation} \label{eq_carre}
    \nabla^2 g(X,h)= \norm{X-h}(\nabla^2 g(X,h))^2.
\end{equation} 
In addition, according to Taylor's theorem,
$$\nabla g(X,h+\alpha Z) - \nabla g(X,h) = \int_{0}^{1} \nabla^2 g(X,h+t\alpha Z) dt\alpha Z,$$
where $Z\sim\mathcal{N}(0,I_p)$ and $\alpha > 0$. Therefore, an estimate of $\nabla^2 G(m)$ is given by
\begin{equation}\label{est_H}
\hat{H}_n= \frac{1}{n+1}\pa{\sum_{k=1}^n\frac{\norm{X_k-m}}{\alpha_k^2}\Phi_k\Phi_k^T+H_0},
\end{equation}
where $\alpha_k = \frac{1}{k\ln{(k+1)}}$, $H_0 = I_p$ and $\Phi_k$ is defined by
\begin{align*}
    \Phi_k &:=  \nabla g(X_k,m+\alpha_k Z_k) - \nabla g(X_k,m)= \int_{0}^{1} \nabla^2 g(X_k,m+t\alpha_k Z_k) dt\alpha_k Z_k,
\end{align*}
where $\pa{Z_k}_k$ are standard independent Gaussian vectors for any $k\ge1$. Indeed, one can check that $\mathbb{E}\cro{\frac{\norm{X_k-m}}{\alpha_k^2}\Phi_k\Phi_k^T} \rightarrow \nabla^2G(m)$. In addition, with the help of Riccati's formula \cite{duflo1997random}, $H_{n+1}^{-1} = (n+1)^{-1}\hat{H}_n^{-1}$ can be easily updated as
$$H_{n+1}^{-1} = H_n^{-1}-\frac{\norm{X_{n+1}-m}}{\alpha_{n+1}^2}\pa{1+\frac{\norm{X_{n+1}-m}}{\alpha_{n+1}^2}\phi_{n+1}^TH_n^{-1}\phi_{n+1}}^{-1}H_n^{-1}\phi_{n+1}\phi_{n+1}^TH_n^{-1}.$$
Thus, knowing $m$, we are able to estimate recursively the inverse of the Hessian with complexity $\mathcal{O}\pa{p^2}$ (instead of  $\mathcal{O}\pa{p^3}$) for each iteration.
\section{Stochastic Newton methods} \label{sec::algos}
In this section we introduce two stochastic Newton methods 
for estimating the median $m$ : 
a stochastic Newton algorithm and its weighted averaged version.
We also give theoretical guarantees on their convergence. 
We recall that $(X_n)_{n\geq 1}$  is a sequence of independent random vectors, of same distribution as vector $X$
and $(Z_n)_{n \geq 1}$ is a sequence of independent standard Gaussian vectors.

\subsection{Stochastic Newton algorithm}
\subsubsection{Definition of the algorithm}\label{subsec::algoSN}
We now introduce stochastic Newton estimates, defined recursively for all $n\ge0$ by
\begin{equation}\label{defSN}
m_{n+1} = m_n + \frac{1}{n+1}\widetilde{H}_n^{-1}\frac{X_{n+1}-m_n}{\norm{X_{n+1}-m_n}},  	
\end{equation}
where $m_0$ is bounded. 
Let $(\tilde{\beta}_n)_{n\geq 1}$ be the strictly positive sequence of real numbers 
defined for any $n\geq 1$ by $\tilde{\beta}_n =\frac{c_{\beta}}{n^{\beta}}$ with $0<\beta<\frac{1}{2}$ and $c_{\beta} >0$. 
The matrix $\widetilde{H}_n$ is given for any $n \geq 0$ by 
\begin{equation} \label{defHntilde}
\widetilde{H}_n = \overline{H}_{n} + \frac{1}{n+1}\sum_{k=1}^n\tilde{\beta}_{k}Z_{k}Z_{k}^T,
\end{equation}
where $\overline{H}_{n}$ is the recursive estimate of the Hessian $\nabla^2G(m)$ defined by 
$$\overline{H}_n = \frac{1}{n+1}\pa{\sum_{k=1}^n\frac{\norm{X_k-m_{k-1}}}{\alpha_k^2}\phi_k\phi_k^T+H_0},$$
with for any $k\geq 1$, 
$\phi_k =  \nabla g(X_k,m_{k-1}+\alpha_k Z_k) - \nabla g(X_k,m_{k-1})$,  
$\alpha_k = \frac{1}{k\ln{(k+1)}}$ and $H_{0}$ is symmetric positive.
We add the term $\sum_{k=1}^n\tilde{\beta}_{k}Z_{k}Z_{k}^T$ 
in order to control the eigenvalues of the Hessian estimate (see Section \ref{proof}), which is necessary to obtain the convergence of the algorithm \cite{boyer2022asymptotic}. 
Thanks to Riccati's formula \cite{duflo1997random} , 
 $H_n^{-1} = (n+1)^{-1}\overline{H}_n^{-1}$ can be updated in two steps, leading to
\begin{align*}
	H_{n+1/2}^{-1} &= H_n^{-1}-\frac{\norm{X_{n+1}-m_{n}}}{\alpha_{n+1}^2}\pa{1+\frac{\norm{X_{n+1}-m_{n}}}{\alpha_{n+1}^2}\phi_{n+1}^TH_n^{-1}\phi_{n+1}}^{-1}H_n^{-1}\phi_{n+1}\phi_{n+1}^TH_n^{-1},\\
	H_{n+1}^{-1} &= H_{n+1/2}^{-1} -\tilde{\beta}_k\pa{1+\tilde{\beta}_kZ_{n+1}^TH_{n+1/2}^{-1}Z_{n+1}}^{-1}H_{n+1/2}^{-1}Z_{n+1}Z_{n+1}^TH_{n+1/2}^{-1}.
\end{align*}
Therefore, this algorithm allows us to update the estimation of the Hessian matrix and the estimation of the geometric median in a recursive way.

\subsubsection{Convergence results}
The following theorem gives the almost sure rates of convergence as well as the asymptotic efficiency of the stochastic Newton estimates. 
Note that its asymptotic efficiency allows us to construct confidence intervals and carry out tests (discussed in Section \ref{ICandTests}).

\begin{theoreme} \label{convergenceNS}
Assume that Assumptions 1 and  2 hold, then the stochastic Newton estimate $m_n$ defined by \eqref{defSN} converges almost surely towards $m$ and 
$$ \norm{m_n-m}^2 = \mathcal{O}\pa{\frac{\ln n}{n}} a.s.$$
Furthermore, the Hessian estimates defined in \eqref{defHntilde} satisfy for all $\delta > 0$
$$\norm{\widetilde{H}_n-H}^2 = \mathcal{O}\pa{\max{\acco{\frac{(\ln n)^{1+\delta}}{n},\frac{c_\beta}{n^{2\beta}}}}}a.s.$$
Finally,
$$\sqrt{n}\pa{m_n-m} \xrightarrow [n\to + \infty] {\mathcal {L}} \mathcal{N}\pa{0,H^{-1}\Sigma H^{-1}},$$
where $\Sigma = \mathbb{E}\cro{\nabla g(X,m)\nabla g(X,m)^T}.$
\end{theoreme}
The proof is given in Section \ref{proof}. Observe that the price to pay in order to control the eigenvalues of the estimates of the Hessian is a loss in term of rate of convergence of the estimates. More precisely, it makes appear a term which converges at a rate $n^{-2\beta}$ instead of $n^{-1}$.

\subsection{Weighted Averaged Stochastic Newton Algorithm}
\subsubsection{Definition of the algorithm}
In order to improve in practice the behavior of the estimates in case of bad initializations, 
we now introduce a Weighted Averaged Stochastic Newton algorithm (WASN) \cite{boyer2022asymptotic} defined recursively for all $n \ge 0$ by:
\begin{align}
    \hat{m}_{n+1} &= \hat{m}_n + \frac{c_\gamma}{\pa{n+1+c'_\gamma}^\gamma}\widetilde{H}_{n,\tau}^{-1}\frac{X_{n+1}-\hat{m}_n}{\norm{X_{n+1}-\hat{m}_n}} \label{defWASN1}\\
    m_{n+1,\tau} &= (1-\tau_{n+1})m_{n,\tau} + \tau_{n+1}\hat{m}_{n+1},\label{defWASN2}
\end{align}
where $c_\gamma >0$, $c'_\gamma \ge 0$ and $\gamma\in\pa{\frac{1}{2},1}$. 
The weighted averaging sequence $(\tau_n)_{n\geq 1}$ is chosen of the following way :  $\tau_n = \frac{\ln (n+1)^\omega}{\sum_{k=0}^n \ln (k+1)^\omega}$ for any $n\ge0$ and $\omega \geq 0$. 
Notice that the case where $\omega = 0$ corresponds to the averaged stochastic Newton algorithm (ASN). 
The recursive estimate of the Hessian   is defined by :
\begin{equation} \label{defHntildeTau}
	\widetilde{H}_{n,\tau} = \frac{1}{n+1}\pa{\sum_{k=1}^n\frac{\norm{X_k-m_{k-1,\tau}}}{\alpha_k^2}\phi_{k,\tau}\phi_{k,\tau}^T+H_0} + \frac{1}{n+1}\sum_{k=1}^n\tilde{\beta}_{k}Z_{k}Z_{k}^T.
\end{equation}
where for any $k\geq 1$, 
$\phi_{k,\tau} =  \nabla g(X_k,m_{k-1,\tau}+\alpha_k Z_k) - \nabla g(X_k,m_{k-1,\tau})$. In order to control the eigenvalue of $\widetilde{H}_{n,\tau}$, $(\tilde{\beta}_n)_{n\geq 1}$ should be the sequence of real numbers defined by 
$\tilde{\beta}_n =\frac{c_{\beta}}{n^{1-\beta}}$ with $0<\beta<\gamma -\frac{1}{2}$ and $c_{\beta} >0$. 
Following the same procedure as for the stochastic Newton algorithm, we can always update $H_{n,\tau}^{-1} = (n+1)^{-1}\overline{H}_{n,\tau}^{-1}$ with Riccati's formula \cite{duflo1997random}.

\subsubsection{Convergence results}
The following theorem shows that under identical assumptions, the WASN estimates are still asymptotically efficient.
\begin{theoreme} \label{convergenceWASN}
Suppose Assumptions 1 and   2 hold, then the Weighted Averaged Stochastic Newton estimates $\hat{m}_n$ and $m_n,\tau$ converge  almost surely towards $m$. In addition,
$$ \norm{\hat{m}_n-m}^2 = \mathcal{O}\pa{\frac{\ln n}{n^\gamma}} a.s.
\qquad \text{and} \qquad
 \norm{m_{n,\tau}-m}^2 = \mathcal{O}\pa{\frac{\ln n}{n}} a.s.$$
Furthermore, the Hessian estimate defined by \eqref{defHntildeTau} satisfies for all $\delta > 0$
$$\norm{\widetilde{H}_{n,\tau}-H}^2 = \mathcal{O}\pa{\max{\acco{\frac{(\ln n)^{1+\delta}}{n},\frac{c_\beta}{n^{2\beta}}}}}a.s.$$
Finally,
$$\sqrt{n}\pa{m_{n,\tau}-m} \xrightarrow [n\to + \infty] {\mathcal {L}} \mathcal{N}\pa{0,H^{-1}\Sigma H^{-1}},$$
where $\Sigma = \mathbb{E}\cro{\nabla g(X,m)\nabla g(X,m)^T}.$
\end{theoreme}
The proof is given in Section \ref{proof}.

\subsection{Comparison of the methods} \label{simuEQM}
We perform a numerical experiment in order to compare the performances of the Stochastic Newton algorithm (SN), the Averaged Stochastic Newton algorithm (ASN), the Weighted Averaged stochastic Newton Algorithm (WASN) and the averaged stochastic gradient descent (ASGD) proposed in \cite{cardot2013efficient}. 
For WASN, we choose $\tau_n = \frac{\ln (n+1)^2}{\sum_{k=0}^n \ln (k+1)^2}$. 
In this experiment, we generate samples of Gaussian random vector $X\sim\mathcal{N}(0_p,\Sigmabar)$ with $p = 10$, and we consider two structures of covariance matrix $\Sigmabar$ defined by
\begin{enumerate}[(i)]
    \item \label{structure1} $\Sigmabar_{ij} = 0.5^{\left|i-j\right|};$
    \item \label{structure2} $\Sigmabar$ is diagonal with $\Sigmabar_{1,1} = 1000 \text{ and } \Sigmabar_{i,i} = 1 \text{ for } i\neq 1.$ 
\end{enumerate}
To evaluate the performances of algorithms, we compute the following mean squared error:
$$MSE(\hat{m}) = \mathbb{E}\cro{\norm{m-\hat{m}}^2},$$
where $\hat{m}$ is an estimate of the median. We estimate this error through Monte-Carlo experiments with $N = 400$ samples, for each sample we generate $n= 15000$ copies of $X$.
In order to see the impact of the initialization of $\hat{m}$, we consider four different initializations : $m_0 = r U$ with $U \sim \mathcal{N}_p\pa{0,I_p}$ and $r = 1$, $5$, $10$ or $15$.
\begin{figure}[H]
		\centering
		\includegraphics[width=0.75\textwidth]{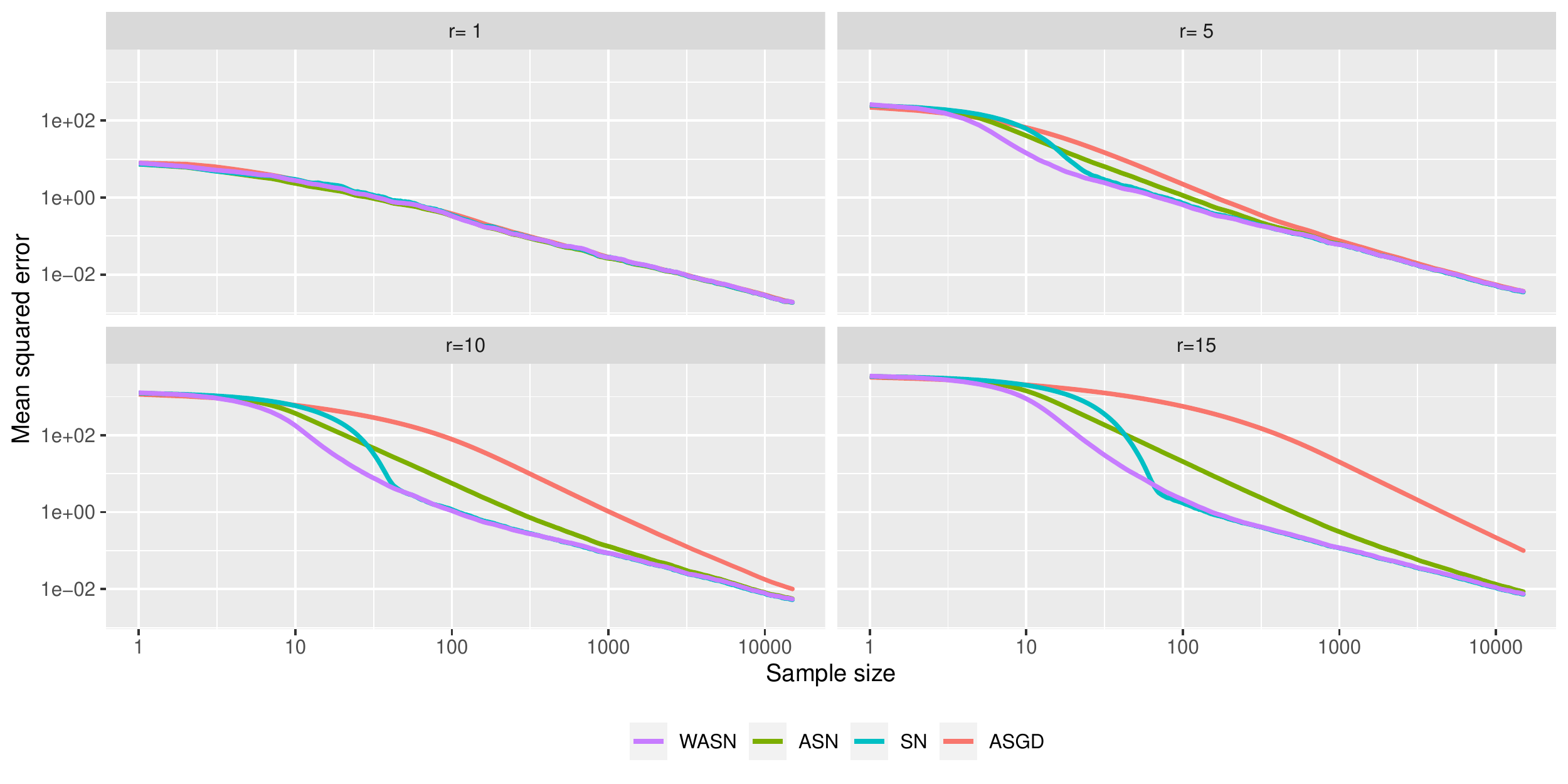}
	\caption{Evolution of the mean squared error with respect to the sample size for structure (\ref{structure1}).}
\end{figure}
Considering the structure (\ref{structure1}), the performances of four algorithms are identical for a good initialization. However, when initialization get worse, we can see that second order methods converge faster than  ASGD.
\begin{figure}[H]
		\centering
		\includegraphics[width=0.75\textwidth]{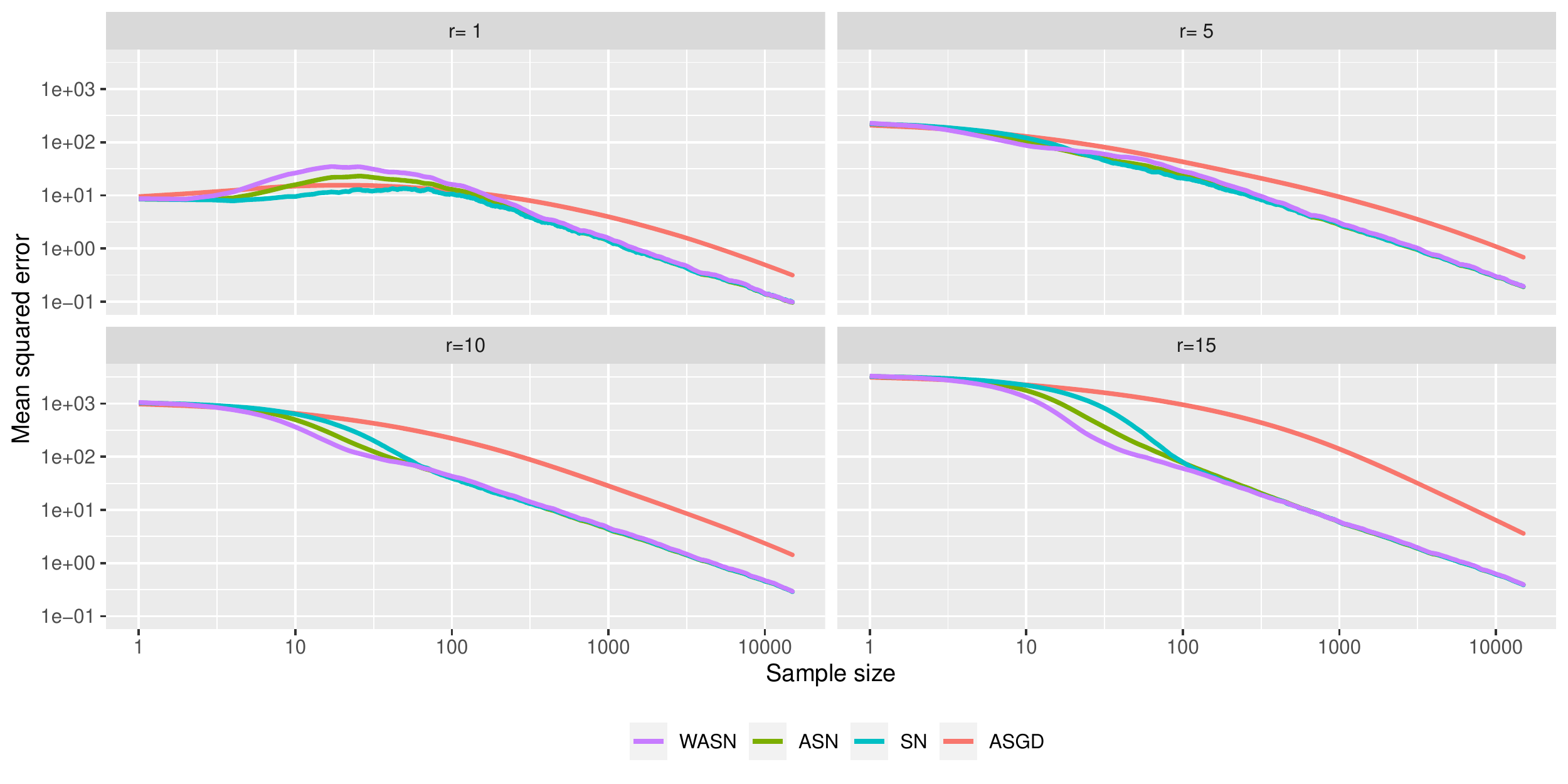}
		\caption{Evolution of the mean squared error with respect to the sample size for structure (\ref{structure2}).}
\end{figure}

When considering the structure (\ref{structure2}), we observe that the Newton type algorithms perform much better than ASGD. Even with a good initialization, the convergence of ASGD is clearly slower than WASN, ASN and SN. Thus ASGD is more sensitive to the structure of the Hessian. Note that for bad initializations, WASN estimators seem to achieve converge faster, and that this phenomenon can be accentuated in the case of even worse conditioned problems, i.e. for even worse Hessian structures \citep{boyer2022asymptotic}.
\section{Confidence intervals and tests} \label{ICandTests}

In this section, we shall propose confidende intervals 
and statistical tests for the median.
These results are obtained from Theorems \ref{convergenceNS} and \ref{convergenceWASN},
and therefore require recursive estimates of the covariance matrix $\Sigma$ defined by
$$\Sigma = \mathbb{E}\cro{\nabla g(X,m)\nabla g(X,m)^T}$$
supposed here positive. 
In the sequel of the section, $\widetilde{m}_n$ will denote any asymptotically efficient estimate of the geometric median.
For example, $\widetilde{m}_n$ can be the ASGD estimate defined by \eqref{defASGD2}, 
or the SN estimate defined by \eqref{defSN}, or the WASN estimate defined by \eqref{defWASN2}. 

\subsection{Estimating the covariance} \label{est_cov} 

A natural recursive estimate of $\Sigma$ is given by 
$$\overline{\Sigma}_n = \frac{1}{n+1} \pa{\sum_{k=1}^n \frac{\pa{X_k-\widetilde{m}_{k-1}}}{\norm{X_k-\widetilde{m}_{k-1}}}\frac{\pa{X_k-\widetilde{m}_{k-1}}^T}{\norm{X_k-\widetilde{m}_{k-1}}} + \Sigma_0},$$
where $\Sigma_0 $ is symmetric positive.
As well as for $H_n$, the Riccati's formula (\cite{duflo1997random}) 
allows us to recursively update matrix $W_{n+1}^{-1} = (n+1)^{-1}\overline{\Sigma}_n^{-1}$ :
\begin{equation} \label{inverseSigma}
	W_{n+1}^{-1} = W_n^{-1}-\pa{1+\frac{\pa{\widetilde{X}_{n+1}}^T}{\norm{\widetilde{X}_{n+1}}}W_n^{-1}\frac{\pa{\widetilde{X}_{n+1}}}{\norm{\widetilde{X}_{n+1}}}}^{-1}W_n^{-1}\frac{\pa{\widetilde{X}_{n+1}}}{\norm{\widetilde{X}_{n+1}}}\frac{\pa{\widetilde{X}_{n+1}}^T}{\norm{\widetilde{X}_{n+1}}}W_n^{-1},
\end{equation}
where $\widetilde{X}_{n+1}:=X_{n+1}-\widetilde{m}_{n}$.
This property will be of particular interest to build online tests (see section \ref{sec::tests}). 
The following theorem gives the rate of convergence of $\overline{\Sigma}_n$.
\begin{theoreme} \label{convergenceSigma}
	Let $\widetilde{m}_n$ be an estimate defined by \eqref{defASGD2}, \eqref{defSN} or \eqref{defWASN2}. Suppose Assumptions 1 and 2 hold, then  for any $\delta > 0$,
	$$ \norm{\overline{\Sigma}_n-\Sigma}^2 = o\pa{\frac{(\ln n)^{1+\delta}}{n}} a.s.$$
\end{theoreme}
The proof is given in section \ref{proof}.

\subsection{Confidence intervals and statistical hypothesis tests}\label{sec::tests}

Let us recall that under Assumptions 1 and 2
$$\sqrt{n}\pa{\widetilde{m}_n-m} \xrightarrow [n\to + \infty] {\mathcal {L}} \mathcal{N}\pa{0,H^{-1}\Sigma H^{-1}}.$$
Thus, we have for any $x_0 \in \mathbb{R}^p\text{\textbackslash}\{0\}$
$$\frac{\sqrt{n}}{\sqrt{x_0^T\overline{S}_n^{-1}\overline{\Sigma}_n \overline{S}_n^{-1}x_0}}\pa{x_0^T\widetilde{m}_n-x_0^Tm}  \xrightarrow [n\to + \infty] {\mathcal {L}} \mathcal{N}\pa{0,1},$$
where 
$$\overline{S}_n = \frac{1}{n+1}\pa{\sum_{k=1}^n\frac{\norm{X_k-\widetilde{m}_{k-1}}}{\alpha_k^2}\widetilde{\phi}_k\widetilde{\phi}_k^T+S_0},$$
with $S_0$ symmetric positive, 
$\pa{Z_k}_k$ standard independent Gaussian vectors, 
and $\widetilde{\phi}_k$ defined by
$\widetilde{\phi}_k =  \nabla g(X_k,\widetilde{m}_{k-1}+\alpha_k Z_k) - \nabla g(X_k,\widetilde{m}_{k-1})$.
As $\overline{S}_n^{-1}$ and $\overline{\Sigma}_n$ can be recursively calculated (see Section \ref{subsec::algoSN} for the update of $\overline{S}_n^{-1}$), 
we can then compute an online confidence interval of $x_0^Tm$, which means that we can determine the confidence interval of the median in any designated direction.
Moreover, since $\widetilde{m}_n$ is asymptotically efficient, one has
$$n\pa{\widetilde{m}_n-m}^T\overline{H}^{*}_n\overline{\Sigma}_n^{-1}\overline{H}^{*}_n\pa{\widetilde{m}_n-m}  \xrightarrow [n\to + \infty] {\mathcal {L}} \mathcal{X}^2_p,$$
where 
$$\overline{H}^{*}_n = \frac{1}{n+1}\pa{\sum_{k=1}^n\frac{1}{\norm{X_k-\widetilde{m}_{k-1}}}\pa{I_p - \frac{(X_k-\widetilde{m}_{k-1})(X_k-\widetilde{m}_{k-1})^T}{\norm{X_k-\widetilde{m}_{k-1}}^2}}+\overline{H}^{*}_0}$$
with $\overline{H}^{*}_0$ symmetric positive. 
Thus $\overline{H}^{*}_n$ can be computed in a recursive way. 
Recall that $\overline{\Sigma}_n^{-1}$ can also be recursively updated with \eqref{inverseSigma}, so that we can perform an online statistical hypothesis test with significance level $\alpha \in (0,1)$  : 
$H_0: "m = m_{test}"$ versus $H_1 : "m\neq m_{test}"$.
We calculate the test statistic $Z_n$ by
$$ Z_n = n\pa{\tilde{m}_n-m_{test}}^T\overline{H}^{*}_n\overline{\Sigma}_n^{-1}\overline{H}^{*}_n\pa{\tilde{m}_n-m_{test}},$$
and we reject the null hypothesis if $Z_n > \zeta_{1-\alpha,p}$, 
where $\zeta_{1-\alpha,p}$ is the quantile of order $1-\alpha$ of the chi-squared distribution with $p$ degrees of freedom. 

\subsection{Simulations}
We now evaluate performances of the different algorithms by studying the empirical levels under $H_0$.
To this aim, we generate samples of size $n = 3000$ of a Gaussian random vector $X\sim\mathcal{N}(0_p,\Sigmabar)$ with $p = 10$, 
where we consider two structures of the covariance matrix $\Sigmabar$ defined in Section \ref{simuEQM}. 
We compute the empirical levels through experiments with $N = 1000$ samples. 
We consider two different initializations :
$m_0 = rU$ with $U \sim \mathcal{N}_p\pa{0,I_p}$ and $r = 1$ or $5$.

\begin{table}[H]
	\begin{center}
		\begin{tabular}{|c|c|c|c|}
			\hline
			Structure of $\Sigmabar$                                                       & $m_0$                          & Algorithm & Empirical level (\%) \\ \hline
			\multirow{8}{*}{(\ref{structure1})}  & \multirow{4}{*}{$U$}   & WASN      & 6.1               \\ \cline{3-4} 
			&                               & ASN       & 5.2               \\ \cline{3-4} 
			&                               & SN        & 5.8               \\ \cline{3-4} 
			&                               & ASGD      & 6.3               \\ \cline{2-4} 
			& \multirow{4}{*}{$5U$}   & WASN      & 5.9               \\ \cline{3-4} 
			&                               & ASN       & 10.7               \\ \cline{3-4} 
			&                               & SN        & 5.4               \\ \cline{3-4} 
			&                               & ASGD      & 23.0                
			\\ \hline
			\multirow{8}{*}{(\ref{structure2})}  & \multirow{4}{*}{$U$}   & WASN      & 18.9               \\ \cline{3-4} 
			&                               & ASN       & 16.8               \\ \cline{3-4} 
			&                               & SN        & 44.3              \\ \cline{3-4} 
			&                               & ASGD      & 56.8              \\ \cline{2-4} 
			& \multirow{4}{*}{$5U$}   & WASN      & 19.1               \\ \cline{3-4} 
			&                               & ASN       & 22.0               \\ \cline{3-4} 
			&                               & SN        & 50.4              \\ \cline{3-4} 
			&                               & ASGD      & 97.2              \\ \hline
			
		\end{tabular}
		\caption{Empirical levels achieved by different algorithms under $H_0$}
		
	\end{center}
\end{table}

The performance of WASN is noticeable, 
it gives lower empirical level and closer to the 5\% theoretical level even if $m_0$ is not well initialized. 
Same as what we observed in previous experiments, 
the algorithm ASN is sensitive to the initializations. 
We can also observe that the empirical level achieved by ASGD is the highest in every considered case. 
In general, proposed second-order methods achieve better results than ASGD, 
and the improvements are more significant in the case where we consider the structure (\ref{structure2}).

In addition, as the statistic test has a chi-squared limit distribution under the null hypothesis, 
we are now interested in the closeness between the simulated distribution of the test statistic and the theoretical distribution. 
For this purpose, we plot the estimated probability densities obtained from different algorithms and the chi-square probability density. 
We can see that with second-order estimates, the estimated distributions are closer to the theoretical distribution, especially with WASN estimates.
\begin{figure}[H]
	\centering
	\includegraphics[width=0.75\textwidth]{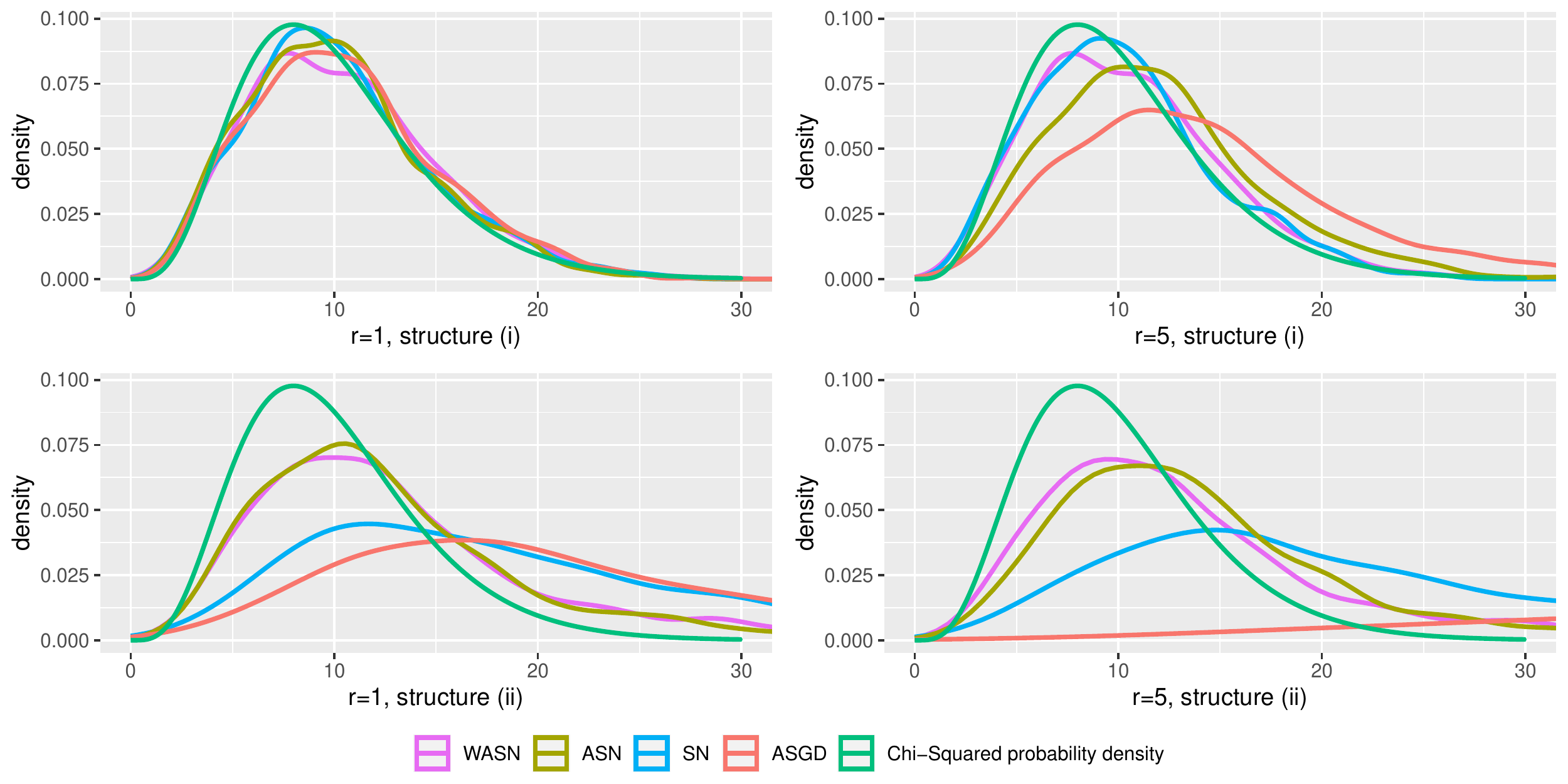}
	\caption{Simulated distributions of the test statistic and the theoretical distribution under $H_0$}
\end{figure}
\section{Proofs} \label{proof}
 In the following, $\norm{\cdot}$ indicates the Euclidean norm for vectors or the spectral norm for matrices.
\subsection{Proof of Theorems \ref{convergenceNS} and \ref{convergenceWASN}}
Remark that the proofs of Theorems \ref{convergenceNS} and \ref{convergenceWASN} are very close. We therefore give the proof of Theorem \ref{convergenceNS} and just highlight differences with the help of remarks. Our objective is to apply Theorem 3.3 (or Theorem 4.3) in \cite{boyer2022asymptotic}. To do so, we are going to verify that the hypotheses given in \cite{boyer2022asymptotic}, termed \textbf{(A1b)}, \textbf{(A1c)}, \textbf{(A2a)}, \textbf{(A2b)}, \textbf{(A2c)}, \textbf{(H1)}, \textbf{(H2a)} and \textbf{(H2b)} are satisfied.
\subsubsection{Verification of conditions on the function to minimize}
First we are going to verify the hypotheses that the function should be met. 
\paragraph{Verification of (A1a).} Assumption 1 ensures that the median $m$ is the unique solution (see \cite{kemperman1987median} and \cite{cardot2013efficient}) of the equation
$$\nabla G(h) = 0,$$
so that Hypothesis \textbf{(A1a)} is satisfied.
\paragraph{Verification of (A1b).} Recall that $\nabla g(X,h) = - \frac{X-h}{\norm{X-h}}$, so that for all $h \in \mathbb{R}^p$,
$$\norm{\nabla g(X,h)} \le 1,$$
Hypothesis \textbf{(A1b)} is then satisfied.
\paragraph{Verification of (A1c).} We have for all $h \in \mathbb{R}^p$
\begin{align*}
    \Sigma(h) &= \mathbb{E}\cro{\nabla g(X,h)\nabla g(X,h)^T} = \mathbb{E}\cro{\frac{(X-h)(X-h)^T)}{\norm{X-h}^2}}.
\end{align*}
The function $\Sigma$ is continuous on $\mathbb{R}^p$, thus Hypothesis \textbf{(A1c)} is satisfied.
\paragraph{Verification of (A2a).} For all $h \in \mathbb{R}^p$,
$$\norm{\nabla^2G(h)} \le \mathbb{E}\cro{\frac{1}{\norm{X-h}}\norm{I_p - \frac{(X-h)(X-h)^T)}{\norm{X-h}^2}}} \le \mathbb{E}\cro{\frac{1}{\norm{X-h}}},$$
and Assumption 1 ensures that
$$\mathbb{E}\cro{\frac{1}{\norm{X-h}}}\le C_6^{\frac{1}{6}},$$
so that Hypothesis \textbf{(A2a)} is satisfied.
\paragraph{Verification of (A2b).} Under Assumption 1 and Assumption 2, for all $h \in \mathbb{R}^p$ the Hessian $\nabla^2 G(h)$ is positive definite (see Section 2.2 in \cite{cardot2013efficient}), thus Hypothesis \textbf{(A2b)} is satisfied.
\paragraph{Verification of (A2c).} Under Assumption 1, the Hessian $\nabla^2 G(h)$ is $6C_6^{\frac{1}{3}}$-Lipschitz (see Lemma \ref{lemma 1}), so that Hypothesis \textbf{(A2c)} is satisfied.

\subsubsection{Controllability of eigenvalues of the Hessian estimator and consistency} \label{controlHessian}
\paragraph{Verification of (H1).} We are going to verify if eigenvalues of the Hessian estimator are well controlled. We recall that $$\widetilde{H}_n = \overline{H}_{n} + \frac{1}{n+1}\sum_{k=1}^n\tilde{\beta}_{k}Z_{k}Z_{k}^T,$$
with $\tilde{\beta}_k =\frac{c_{\beta}}{k^{\beta}}$ where $0<\beta<\frac{1}{2}$ and $c_{\beta} >0$. 
\begin{rmq}
	For WASN the condition on $\beta$ should be $\beta<\gamma - \frac{1}{2}$ instead of $\beta<\frac{1}{2}$.
\end{rmq}
Thus we have
$$\lambda_{min}(\tilde{H_n}) \ge \frac{\lambda_{min}\pa{H_0}}{n+1} + \frac{1}{n+1}\lambda_{min}\pa{\sum_{k=1}^n\tilde{\beta}_{k}Z_{k}Z_{k}^T}.$$
In addition, 
$$\pa{\frac{c_\beta}{1-\beta}n^{1-\beta}}^{-1}\sum_{k=1}^n\tilde{\beta}_{k}Z_{k}Z_{k}^T \xrightarrow[n\to+\infty]{a.s.} I_p,$$
so that $\lambda_{max}\pa{\widetilde{H}_n^{-1}} = \mathcal{O}\pa{n^\beta}$.
For the largest eigenvalue of $\overline{H}_n$, we have
\begin{align*}
    \norm{\overline{H}_n}  &\le \frac{1}{n+1}\sum^n_{k=1} \frac{\norm{X_k-m_{k-1}}}{\alpha_k^2}\norm{\phi_k}^2 + \frac{1}{n+1}\norm{H_0}\\
    &=  \frac{1}{n+1}\sum^n_{k=1} \frac{\norm{X_k-m_{k-1}}}{\alpha_k^2}\norm{\nabla g(X_k,m_{k-1}) - \nabla g(X_k,m_{k-1}+\alpha_kZ_k)}^2 + \frac{1}{n+1}\norm{H_0} \\
    &= \frac{1}{n+1}\sum^n_{k=1} \frac{\norm{X_k-m_{k-1}}}{\alpha_k^2}\norm{\frac{X_k-m_{k-1}}{\norm{X_k-m_{k-1}}}-\frac{X_k-\pa{m_{k-1}+\alpha_kZ_k}}{\norm{X_k-\pa{m_{k-1}+\alpha_kZ_k}}}}^2 + \frac{1}{n+1}\norm{H_0}.\\
\end{align*}
Since $\norm{\frac{C-A}{\norm{C-A}}-\frac{B-A}{\norm{B-A}}} \le 2\frac{\norm{C-B}}{\norm{B-A}}$ (see \cite{cardot2012fast} page 25), we have
\begin{align*}
    \norm{\overline{H}_n}  &\le \frac{4}{n+1}\sum^n_{k=1} \frac{\norm{X_k-m_{k-1}}}{\alpha_k^2}\norm{\frac{\alpha_kZ_k}{\norm{X_k-m_{k-1}}}}^2 +\frac{1}{n+1}\norm{H_0} \\
    &= \frac{1}{n+1}\sum^n_{k=1} \frac{4\norm{Z_k}^2}{\norm{X_k-m_{k-1}}} + \frac{1}{n+1}\norm{H_0}.
\end{align*}
Thanks to Assumption 1, by independence between $Z_k$ and $X_k$ and by Hölder's inequality, we have $$\mathbb{E}\cro{\frac{4\norm{Z_k}^2}{\norm{X_k-m_{k-1}}}\,|\,\mathcal{F}_{k-1}} = 4\mathbb{E}\cro{\norm{Z_k}^2}\mathbb{E}\cro{\norm{X_k-m_{k-1}}^{-1}}\le 4p C_6^{\frac{1}{6}},$$ so that $$\frac{1}{n+1}\sum_{k=1}^n\mathbb{E}\cro{\frac{4\norm{Z_k}^2}{\norm{X_k-m_{k-1}}}\,|\,\mathcal{F}_{k-1}}\le 4p C_6^{\frac{1}{6}}.$$
Moreover, with analogous calculs, one has
$$\mathbb{E}\cro{\pa{\frac{4\norm{Z_k}^2}{\norm{X_k-m_{k-1}}}}^2\,|\,\mathcal{F}_{k-1}} = 16\mathbb{E}\cro{\norm{Z_k}^4}\mathbb{E}\cro{\norm{X_k-m_{k-1}}^{-2}}\le 16p(p+2)C_6^{\frac{1}{3}},$$
With the help of law of large numbers for martingales, for all $\delta > 0$,
$$\pa{\frac{1}{n}\sum_{k=1}^n\frac{4\norm{Z_k}^2}{\norm{X_k-m_{k-1}}} -\mathbb{E}\cro{\frac{4\norm{Z_k}^2}{\norm{X_k-m_{k-1}}}\,|\,\mathcal{F}_{k-1}}}^2  = o\pa{\frac{(\ln n)^{1+\delta}}{n}}  a.s. $$
Thus, 
\begin{small}
\begin{align*}
\norm{\overline{H}_n}&\le\norm{\frac{1}{n+1}\sum_{k=1}^n\mathbb{E}\cro{\frac{4\norm{Z_k}^2}{\norm{X_k-m_{k-1}}}\,|\,\mathcal{F}_{k-1}}} + \norm{\frac{1}{n+1}\sum_{k=1}^n\frac{4\norm{Z_k}^2}{\norm{X_k-m_{k-1}}}-\mathbb{E}\cro{\frac{4\norm{Z_k}^2}{\norm{X_k-m_{k-1}}}\,|\,\mathcal{F}_{k-1}}} \\
&\qquad +\frac{1}{n+1}\norm{H_0} \\
&= \mathcal{O}(1) \text{ } a.s.
\end{align*}
\end{small}
Then, 
\begin{align*}
    \norm{\widetilde{H}_n} &\le \norm{\overline{H}_n} + \frac{1}{n+1}\sum_{k=1}^n\tilde{\beta}_{k}  = \mathcal{O}(1) \text{ } a.s.
\end{align*}
The largest eigenvalue of $\widetilde{H}_n^{-1}$ and $\widetilde{H}_n$ can be controlled, according to Theorem 3.1 in \cite{boyer2022asymptotic},  the stochastic Newton estimator satisfies 
$$m_n\xrightarrow[n\to+\infty]{a.s.}m.$$
\begin{rmq}
	For WASN, according to Theorem 4.1 in \cite{boyer2022asymptotic}, the estimator $\hat{m}_n$ converges almost surely to $m$, which implies the almost sure convergence of $m_{n,\tau}$.
\end{rmq}
\subsubsection{Convergence of the Hessian estimator and rate of convergence}
\paragraph{Verification of (H2a).}We verify now if the Hessian estimator converges towards $\nabla^2G(m)$. We define
$$X_{k,t} := X_k - (m_{k-1}+t\alpha_kZ_k),$$
and
$$w_{k,t} := \frac{1}{\norm{X_{k,t}}}\pa{I_p -  \frac{X_{k,t}X_{k,t}^T}{\norm{X_{k,t}}^2}}.$$
We then have 
\begin{align*}
    \overline{H}_n &= \frac{1}{n+1}\sum_{k=1}^n\frac{\norm{X_k-m_{k-1}}}{\alpha_k^2}\int_0^1w_{k,t}dt\alpha_kZ_k\alpha_kZ_k^T\int_0^1w_{k,t}dt + \frac{1}{n+1}H_0 \\
    &= \frac{1}{n+1}\sum_{k=1}^n\norm{X_k-m_{k-1}}\int_0^1w_{k,t}dtZ_kZ_k^T\int_0^1w_{k,t}dt + \frac{1}{n+1}H_0 \\
    &= \overbrace{\frac{1}{n+1}\sum_{k=1}^n\norm{X_k-m_{k-1}}\int_0^1w_{k,t}-w_{k,0}dtZ_kZ_k^T\int_0^1w_{k,t}dt}^{\mathcal{M}_{1,n}}\\
    &\text{ }+ \overbrace{\frac{1}{n+1}\sum_{k=1}^n\norm{X_k-m_{k-1}}w_{k,0}Z_kZ_k^T\int_0^1w_{k,t}-w_{k,0}dt}^{\mathcal{M}_{2,n}}\\
    &\text{ }+ \overbrace{\frac{1}{n+1}\sum_{k=1}^n\norm{X_k-m_{k-1}}w_{k,0}Z_kZ_k^Tw_{k,0}}^{\mathcal{M}_{3,n}} + \frac{1}{n+1}H_0
\end{align*}
\paragraph{Convergence of $\mathcal{M}_{3,n}$.} We define 
$$\mathcal{Y}_k : = \norm{X_k-m_{k-1}}w_{k,0}Z_kZ_k^Tw_{k,0},$$
remark that one has
$$\mathcal{M}_{3,n} = \frac{1}{n+1}\sum_{k=1}^n\mathcal{Y}_k = \frac{1}{n+1}\sum_{k=1}^n \mathbb{E}\cro{\mathcal{Y}_k\,|\,\mathcal{F}_{k-1}} + \frac{1}{n+1}\sum_{k=1}^n\mathcal{Y}_k-\mathbb{E}\cro{\mathcal{Y}_k\,|\,\mathcal{F}_{k-1}}.$$
First we prove that
$$\frac{1}{n+1}\sum_{k=1}^n\mathbb{E}\cro{\mathcal{Y}_k\,|\,\mathcal{F}_{k-1}}  \xrightarrow[n\to+\infty]{a.s.} \nabla^2G(m).$$
We have
$$w_{k,0} := \frac{1}{\norm{X_k-m_{k-1}}}\pa{I_p - \pa{ \frac{X_k-m_{k-1}}{\norm{X_k-m_{k-1}}}}\pa{ \frac{X_k-m_{k-1}}{\norm{X_k-m_{k-1}}}}^T} = \nabla^2g(X_k,m_{k-1}),$$
so that by equation \eqref{eq_carre}
$$\norm{X_k-m_{k-1}}w_{k,0}^2 = w_{k,0}.$$
In addition, as the estimator of the median satisfies
$$m_n\xrightarrow[n\to+\infty]{a.s.}m,$$
we have by continuity
$$\mathbb{E}\cro{w_{n,0}\,|\,\mathcal{F}_{n-1}}= \nabla^2G(m_n)\xrightarrow[n\to+\infty]{a.s.} \nabla^2G(m).$$
Therefore, as $\{Z_k\}_k$ are standard independent Gaussian vectors, by law of large numbers, we have 
\begin{align*}
\frac{1}{n+1}\sum_{k=1}^n\mathcal\mathbb{E}\cro{\mathcal{Y}_k\,|\,\mathcal{F}_{k-1}} &=\frac{1}{n+1}\sum_{k=1}^n\mathcal\mathbb{E}\cro{\norm{X_k-m_{k-1}}w_{k,0}^2\,|\,\mathcal{F}_{k-1}}\\
&= \frac{1}{n+1}\sum_{k=1}^n\mathcal\mathbb{E}\cro{ w_{k,0}\,|\,\mathcal{F}_{k-1}} \xrightarrow[n\to+\infty]{a.s.} \nabla^2G(m).
\end{align*}
Moreover, thanks to Assumption 1 and by independence, 
\begin{align*}
    \mathbb{E}\cro{\norm{\mathcal{Y}_k}^2\,|\,\mathcal{F}_{k-1}} &\le \mathbb{E}\cro{\norm{X_k-m_{k-1}}^2\norm{w_{k,0}}^4\norm{Z_k}^4\,|\,\mathcal{F}_{k-1}} \\
    &\le \mathbb{E}\cro{\norm{\mathcal{Z}_k}^4\,|\,\mathcal{F}_{k-1}}\mathbb{E}\cro{\norm{\frac{1}{X_k-m_{k-1}}}^2\,|\,\mathcal{F}_{k-1}} \\ 
    &\le p(p+2)C_6^{\frac{1}{3}},
\end{align*}
which results in, with the help of law of large numbers for martingales, that for all $\delta > 0$,
$$\norm{\frac{1}{n}\sum_{k=1}^n\mathcal{Y}_k -\mathbb{E}\cro{\mathcal{Y}_k\,|\,\mathcal{F}_{k-1}}}^2  = o\pa{\frac{(\ln n)^{1+\delta}}{n}}. $$
Thus,
$$ \mathcal{M}_{3,n} \xrightarrow[n\to+\infty]{a.s.} \nabla^2G(m).$$
\paragraph{Convergence of $\mathcal{M}_{2,n}$} : In order to get the rate of convergence of $\mathcal{M}_{2,n}$, let us first introduce a generalization of Lemma 5.1 in \cite{cardot2017online}.
\begin{lemma}\label{lemma 1}
For all  $h, h' \in \mathbb{R}^p$ and $0<q \leq 3$,
$$\left( \mathbb{E}\left[ \norm{\nabla^2g(X,h') - \nabla^2g(X,h)}^{q} \right]\right)^{\frac{1}{q}} \le {6C_{6}^{\frac{1}{3}} \norm{h'-h}}.$$
\end{lemma}
In our case, for all $t \in (0,1)$ and $q \in (0,3]$, and considering the filtration $\mathcal{F}_{k}' = \sigma \left( X_{1} , \ldots ,X_{k-1}, Z_{1}, \ldots ,Z_{k} \right)$, we have
$$ \mathbb{E}\left[ \norm{w_{k,t}-w_{k,0}}^{q} |\mathcal{F}_{k} \right]  \le 6^{q} C_{6}^{\frac{q}{3}}\alpha_k^{q}  \left\| Z_{k} \right\|^{q}.$$
We define 
$$\mathcal{W}_k : = \norm{X_k-m_{k-1}}w_{k,0}Z_kZ_k^T\int_0^1w_{k,t}-w_{k,0}dt.$$
Then,
$$\norm{\mathcal{M}_{2,n}}\le\frac{1}{n+1}\sum_{k=1}^n\mathbb{E}\cro{\norm{\mathcal{W}_k}\,|\,\mathcal{F}_{k-1}} + \frac{1}{n+1}\sum_{k=1}^n \norm{\mathcal{W}_k  }-\mathbb{E}\cro{\norm{\mathcal{W}_k}\,|\,\mathcal{F}_{k-1}} .$$
Remark that
\begin{align*}
    \mathbb{E}\cro{\norm{\mathcal{W}_k} \,|\,\mathcal{F}_{k-1}} &\le \mathbb{E}\cro{\norm{X_k-m_{k-1}}\norm{w_{k,0}}\norm{Z_k}^2\norm{\int_0^1w_{k,t}-w_{k,0}}dt\,|\,\mathcal{F}_{k-1}} \\
    &\le \mathbb{E}\cro{\norm{Z_k}^2\int_0^1 \left\| w_{k,t} - w_{k,0} \right\| dt\,|\,\mathcal{F}_{k-1}} \\
    &=  \mathbb{E}\cro{ \norm{Z_k}^2\int_0^1 \mathbb{E} \left[ \left\| w_{k,t} - w_{k,0} \right\|   dt\,|\,\mathcal{F}_{k}' \right]\,|\,\mathcal{F}_{k-1}},
\end{align*}
where $\mathcal{F}_{k'} = \sigma\{X_1,...,X_{k-1},Z_1,...,Z_k\}$. Therefore, thanks to Lemma \ref{lemma 1},  we have
$$ \mathbb{E}\cro{\norm{\mathcal{W}_k} \,|\,\mathcal{F}_{k-1}} \le 6\alpha_k C_6^{\frac{1}{3}} \mathbb{E}\cro{\norm{Z_k}^3 \,|\,\mathcal{F}_{k-1}}.$$
Since $\alpha_k = \frac{1}{k\ln{k+1}}$, it comes $\mathbb{E}\cro{\norm{\mathcal{W}_k} \,|\,\mathcal{F}_{k-1}} = \mathcal{O}\pa{\frac{1}{k\ln{k}}}$, which leads to
$$\frac{1}{n+1}\sum_{k=1}^n\mathbb{E}\cro{\norm{\mathcal{W}_k}\,|\,\mathcal{F}_{k-1}} = \mathcal{O}\pa{\frac{\ln{n}}{n}} a.s.$$
In addition, according to Lemma \ref{lemma 1},  and with the help of Hölder's inequality,
\begin{align*}
     \mathbb{E}\cro{\norm{\mathcal{W}_k}^2 \,|\,\mathcal{F}_{k-1}} &\le \mathbb{E}\cro{ \norm{Z_k}^4 \left( \int_0^1 \left\| w_{k,t} - w_{k,0} \right\| dt \right)^{2}\,|\,\mathcal{F}_{k-1}}^2 \\
     &\le  \mathbb{E}\cro{\mathbb{E}\cro{\norm{Z_k}^4\pa{\int_0^1 \left\| w_{k,t} - w_{k,0} \right\| dt}^2\,|\,\mathcal{F}_{k}'}\,|\,\mathcal{F}_{k-1}} \\
     & \leq \mathbb{E}\left[ \left\| Z_{k} \right\|^{4} \int_{0}^{1} \mathbb{E}\left[ \left\| w_{k,t} - w_{k,0} \right\|^{2} |\mathcal{F}_{k}' \right] dt |\mathcal{F}_{k-1} \right] \\
     &\le 36\alpha_k^2 \mathbb{E}\cro{\norm{Z_k}^6 \,|\,\mathcal{F}_{k-1}}C_6^{\frac{2}{3}} . \\
\end{align*}
Thus, with the help of law of large numbers for martingales,
$$\norm{\frac{1}{n}\sum_{k=1}^n\mathcal{W}_k -\mathbb{E}\cro{\norm{\mathcal{W}_k}\,|\,\mathcal{F}_{k-1}}}^2  = o\pa{\frac{(\ln n)^{1+\delta}}{n}}a.s.$$
Therefore, we obtain
$$ \mathcal{M}_{2,n} = o\pa{\frac{(\ln n)^{1+\delta}}{n}}a.s.$$
\paragraph{Convergence of $\mathcal{M}_{1,n}$.} We define 
$$\mathcal{V}_k : =\norm{X_k-m_{k-1}}\int_0^1w_{k,t}-w_{k,0}dtZ_kZ_k^T\int_0^1w_{k,t}dt.$$
Remark that
\begin{align*}
    \mathbb{E}\cro{\norm{\mathcal{V}_k} \,|\,\mathcal{F}_{k-1}} &\le \mathbb{E}\cro{\norm{X_k-m_{k-1}}\norm{\int_0^1w_{k,t}dt}\norm{Z_k}^2\norm{\int_0^1w_{k,t}-w_{k,0}dt}\,|\,\mathcal{F}_{k-1}} \\
    &\le \mathbb{E}\cro{\norm{Z_k}^2\int_0^1\frac{\norm{X_{k,t}}+t\alpha_k\norm{Z_k}}{\norm{X_{k,t}}}dt\int_0^1 \left\| w_{k,t} - w_{k,0} \right\| dt\,|\,\mathcal{F}_{k-1}} \\
    &\le \mathbb{E}\cro{ \mathbb{E}\left[  \norm{Z_k}^2 \pa{1+\int_0^1\frac{\alpha_k\norm{Z_k}}{\norm{X_{k,t}}}dt}\int_0^1 \left\| w_{k,t} - w_{k,0} \right\|dt\,|\,\mathcal{F}_{k}' \right]\,|\,\mathcal{F}_{k-1}}
\end{align*}
where $\mathcal{F}_{k}' = \sigma\left( X_1,...,X_{k-1},Z_1,...,Z_k \right)$. Thus, according to Lemma \ref{lemma 1} and Assumption 1,  one has with the help of Hölder's inequality
\begin{align*} 
\mathbb{E}\cro{\norm{\mathcal{V}_k} \,|\,\mathcal{F}_{k-1}} &  \le \mathbb{E}\left[ \left\| Z_{k} \right\|^{2} \int_{0}^{1} \mathbb{E}\left[ \left\| w_{k,t} - w_{k,0} \right\| |\mathcal{F}_{k}' \right] dt \right] \\
&  + \mathbb{E}\left[ \alpha_{k}\left\| Z_{k} \right\|^{3} \left( \int_{0}^{1} \mathbb{E}\left[ \frac{1}{\left\| X_{k,t} \right\|^{2}} |\mathcal{F}_{k}' \right] dt \right)^{\frac{1}{2}} \left( \int_{0}^{1} \mathbb{E}\left[ \left\| w_{k,t} - w_{k,0} \right\|^{2} |\mathcal{F}_{k}'  \right] dt \right)^{\frac{1}{2} }   |\mathcal{F}_{k-1} \right]  \\
& \leq 6\alpha_k C_6^{\frac{1}{3}} \mathbb{E}\cro{\norm{Z_k}^3 \,|\,\mathcal{F}_{k-1}} +6\alpha_k^2 \mathbb{E}\left[ \left\| Z_{k} \right\|^{4} |\mathcal{F}_{k-1} \right] C_6^{\frac{1}{2}}.
\end{align*}
We have $\alpha_k = \frac{1}{k\ln{k+1}}$, so that $\mathbb{E}\cro{\norm{\mathcal{V}_k} \,|\,\mathcal{F}_{k-1}} = \mathcal{O}\pa{\frac{1}{k\ln{k}}}$, which leads to
$$\frac{1}{n+1}\sum_{k=1}^n\mathbb{E}\cro{\norm{\mathcal{V}_k}\,|\,\mathcal{F}_{k-1}} = \mathcal{O}\pa{\frac{\ln{n}}{n}}.$$
Furthermore, we have by Hölder's inequality
\begin{align*}
     \mathbb{E}& \cro{\norm{\mathcal{V}_k}^2 \,|\,\mathcal{F}_{k-1}} \le \mathbb{E}\cro{\norm{Z_k}^4 \pa{1+\int_0^1\frac{\alpha_k\norm{Z_k}}{\norm{X_{k,t}}}dt}^2\left( \int_0^1 \left\| w_{k,t} - w_{k,0} \right\| dt\right)^{2}\,|\,\mathcal{F}_{k'}\,|\,\mathcal{F}_{k-1}} \\
     &\le 2 \mathbb{E} \cro{ \left\| Z_{k} \right\|^{4}  \int_0^1 \left\| w_{k,t} - w_{k,0} \right\|^{2}dt  \,|\,\mathcal{F}_{k-1}}+2\alpha_k^2\mathbb{E}\cro{\norm{Z_k}^{6}\int_0^1\int_{0}^{1}\frac{1}{\norm{X_{k,t}}^2}  \left\| w_{k,t'} - w_{k,0} \right\|^{2}dtdt'  \,|\,\mathcal{F}_{k-1}} 
     =:(*)
\end{align*}
and we therefore have, applying Hölder's inequality,
\begin{align*}
(*) &\le 2 \mathbb{E} \cro{ \left\| Z_{k} \right\|^{4}\int_0^1 \mathbb{E} \left[ \left\| w_{k,t} - w_{k,0} \right\|^{2} |\mathcal{F}_{k}' \right]	dt\,|\,\mathcal{F}_{k-1}} \\
& +2\alpha_k^2\mathbb{E}\cro{\norm{Z_k}^{6}   \int_0^1 \int_{0}^{1} \left( \mathbb{E}\left[  \frac{1}{\norm{X_{k,t}}^6} |\mathcal{F}_{k}' \right] \right)^{\frac{1}{3}}  \left( \mathbb{E}\left[  \left\| w_{k,t'} - w_{k,0} \right\|^{3} |\mathcal{F}_{k}' \right]   \right)^{\frac{2}{3}} dt dt' \,|\,\mathcal{F}_{k-1}} .
\end{align*}
Then, thanks to Assumption 1 and Lemma \ref{lemma 1}, 
$$\mathbb{E}\cro{\norm{\mathcal{V}_k}^2 \,|\,\mathcal{F}_{k-1}} \le 72\alpha_k^2C_6^{2/3}\mathbb{E}\cro{\norm{Z_k}^6}+72\alpha_k^4C_6 \mathbb{E}\cro{\norm{Z_k}^8}.$$
With the help of law of large numbers for martingales, one then has
$$\norm{\frac{1}{n}\sum_{k=1}^n\mathcal{V}_k -\mathbb{E}\cro{\mathcal{V}_k\,|\,\mathcal{F}_{k-1}}}^2  = o\pa{\frac{(\ln n)^{1+\delta}}{n}} a.s $$
and
$$ \mathcal{M}_{1,n} = o\pa{\frac{(\ln n)^{1+\delta}}{n}} a.s.$$
Finally, we have 
$$ \overline{H}_n \xrightarrow[n\to+\infty]{a.s.} \nabla^2G(m).$$
Notice that $$\widetilde{H}_n = \overline{H}_{n} + \frac{1}{n+1}\sum_{k=1}^n\tilde{\beta}_{k}Z_{k}Z_{k}^T,$$
and 
$$\frac{1}{n+1}\sum_{k=1}^n\tilde{\beta}_{k}Z_{k}Z_{k}^T \xrightarrow[n\to+\infty]{a.s.} 0.$$
Therefore,   
$$ \widetilde{H}_n \xrightarrow[n\to+\infty]{a.s.} \nabla^2G(m).$$
According to Theorem 3.2 in \cite{boyer2022asymptotic}, the stochastic Newton estimator satisfies
$$ \norm{m_n-m}^2 = \mathcal{O}\pa{\frac{\ln n}{n}} a.s.$$
\begin{rmq}
	For WASN, according to Theorem 4.2 in \cite{boyer2022asymptotic}, we have 
	$$ \norm{\hat{m}_n-m}^2 = \mathcal{O}\pa{\frac{\ln n}{n^\gamma}} a.s., \quad \text{which implies that} \quad \norm{m_{n,\tau}-m}^2 = \mathcal{O}\pa{\frac{\ln n}{n^\gamma}} a.s.$$
\end{rmq}
\subsubsection{Rate of convergence of the Hessian estimator and asymptotic efficiency}
We now give the rate of convergence of $\widetilde{H}_n$. We recall that 
$$w_{k,0} := \frac{1}{\norm{X_k-m_{k-1}}}\pa{I_p - \pa{ \frac{X_k-m_{k-1}}{\norm{X_k-m_{k-1}}}}\pa{ \frac{X_k-m_{k-1}}{\norm{X_k-m_{k-1}}}}^T},$$
which means that
$$\mathbb{E}\cro{w_{k,0}\,|\,\mathcal{F}_{k-1}} = \nabla^2G(m_{k-1}).$$
Note that $\nabla^2G(h)$ is $6C_6^{\frac{1}{3}}$-Lipschitz, so that we have
$$\mathbb{E}\cro{\norm{w_{k,0}-\nabla^2G(m)}^2\,|\,\mathcal{F}_{k-1}} = \norm{\nabla^2G(m_{k-1})-\nabla^2G(m)}^2 \le 6C_6^{\frac{1}{3}} \norm{m_{k-1}-m}^2. $$
As the estimator satisfies
$$\norm{m_n-m}^2 = \mathcal{O}\pa{\frac{\ln n}{n}} a.s.,$$
we have
$$\norm{\mathbb{E}\cro{w_{k,0}\,|\,\mathcal{F}_{k-1}}-\nabla^2G(m)}^2 = \mathcal{O}\pa{\frac{\ln k}{k}} a.s.$$
\begin{rmq}
For WASN, here we have $\norm{m_{n,\tau}-m}^2 = \mathcal{O}\pa{\frac{\ln n}{n^\gamma}} a.s.$, so that\\ $\norm{\mathbb{E}\cro{w_{k,0}\,|\,\mathcal{F}_{k-1}}-\nabla^2G(m)}^2 = \mathcal{O}\pa{\frac{\ln k}{k^\gamma}} a.s.$
\end{rmq}
Therefore, as $\{Z_k\}_k$ are standard independent Gaussian vectors, we have 
\begin{align*}
\norm{\frac{1}{n+1}\sum_{k=1}^n\mathcal\mathbb{E}\cro{\mathcal{Y}_k\,|\,\mathcal{F}_{k-1}} - \nabla^2G(m)}^2 = \mathcal{O}\pa{\frac{\ln n}{n}} a.s.
\end{align*}
We have proved that for all $\delta > 0$
$$\norm{\frac{1}{n}\sum_{k=1}^n\mathcal{Y}_k -\mathbb{E}\cro{\mathcal{Y}_k\,|\,\mathcal{F}_{k-1}}}^2  = o\pa{\frac{(\ln n)^{1+\delta}}{n}}a.s., $$
so that 
$$\norm{\mathcal{M}_{3,n}-\nabla^2G(m)}^2 = o\pa{\frac{(\ln n)^{1+\delta}}{n}}a.s.$$
Finally, the Hessian estimator satisfies for all $\delta > 0$
$$\norm{\widetilde{H}_n-H}^2 = \mathcal{O}\pa{\max{\acco{\frac{(\ln n)^{1+\delta}}{n},\frac{c_\beta}{n^{2\beta}}}}}a.s.$$
According to Theorem 3.3 in \cite{boyer2022asymptotic}, the stochastic Newton estimator satisfies
$$\sqrt{n}\pa{m_n-m} \sim \mathcal{N}\pa{0,H^{-1}\Sigma H^{-1}},$$
where $\Sigma = \mathbb{E}\cro{\nabla g(X,m)\nabla g(X,m)^T}.$
\begin{rmq}
	For WASN, we have first for all $\delta > 0$
	$$\norm{\widetilde{H}_{n,\tau}-H}^2 = \mathcal{O}\pa{\max{\acco{\frac{(\ln n)^{1+\delta}}{n^\gamma},\frac{c_\beta}{n^{2\beta}}}}}a.s.$$
	Then according to Theorem 4.3 in \cite{boyer2022asymptotic}, we have
	$$\norm{m_{n,\tau}-m}^2 = \mathcal{O}\pa{\frac{\ln n}{n}} a.s.,$$
	which results in 
	$$\norm{\widetilde{H}_{n,\tau}-H}^2 = \mathcal{O}\pa{\max{\acco{\frac{(\ln n)^{1+\delta}}{n},\frac{c_\beta}{n^{2\beta}}}}}a.s.
	\quad \text{and} \quad
	\sqrt{n}\pa{m_{n,\tau}-m} \sim \mathcal{N}\pa{0,H^{-1}\Sigma H^{-1}}.$$
\end{rmq}
\subsection{Proof of Theorem \ref{convergenceSigma}}
We define 
$$\mathcal{T}_k : = \frac{\pa{X_k-\widetilde{m}_{k-1}}}{\norm{X_k-\widetilde{m}_{k-1}}}\frac{\pa{X_k-\widetilde{m}_{k-1}}^T}{\norm{X_k-\widetilde{m}_{k-1}}},$$
then one has
$$\Sigma_n = \frac{1}{n+1}\sum_{k=1}^n\mathcal{T}_k + \frac{1}{n+1}\Sigma_0 = \frac{1}{n+1}\sum_{k=1}^n \mathbb{E}\cro{\mathcal{T}_k\,|\,\mathcal{F}_{k-1}} + \frac{1}{n+1}\sum_{k=1}^n\mathcal{T}_k-\mathbb{E}\cro{\mathcal{T}_k\,|\,\mathcal{F}_{k-1}}+ \frac{1}{n+1}\Sigma_0.$$
Note that $\mathbb{E}\cro{\mathcal{T}_k\,|\,\mathcal{F}_{k-1}} = \Sigma(\widetilde{m}_{k-1}).$ In addition, we have 
$$\Sigma(h) = \mathbb{E}\cro{\nabla g(X,h)\nabla g(X,h)^T}$$
Thus, thanks to Hypothesis \textbf{(A2c)}, $\Sigma(h)$ is $6C_6^{\frac{1}{6}}$-Lipschitz (see \cite{godichon2019online} section 6.2), which means
$$\norm{\mathbb{E}\cro{\mathcal{T}_k\,|\,\mathcal{F}_{k-1}} - \Sigma}^2 = \norm{\Sigma(\widetilde{m}_{k-1}) - \Sigma(m)}^2 \le 6C_6^{\frac{1}{6}}\norm{\widetilde{m}_{k-1}-m}^2.$$
As the estimator $\widetilde{m}_k$ satisfies 
$$ \norm{\widetilde{m}_k-m}^2 = \mathcal{O}\pa{\frac{\ln n}{n}} a.s.,$$
we obtain
$$\norm{\mathbb{E}\cro{\mathcal{T}_k\,|\,\mathcal{F}_{k-1}} - \Sigma}^2 = \mathcal{O}\pa{\frac{\ln n}{n}} a.s.$$
Moreover, it is obvious that
$$\mathbb{E}\cro{\norm{\mathcal{T}_k}^2\,|\,\mathcal{F}_{k-1}} \le 1,$$
which leads to, with the help of law of large numbers for martingales,
$$\norm{\frac{1}{n}\sum_{k=1}^n\mathcal{T}_k -\mathbb{E}\cro{\mathcal{T}_k\,|\,\mathcal{F}_{k-1}}}^2  = o\pa{\frac{(\ln n)^{1+\delta}}{n}}a.s. $$
Finally we have
\begin{align*}
\norm{\Sigma_n-\Sigma}^2 &\le \norm{\frac{1}{n+1}\sum_{k=1}^n\mathcal{T}_k -\mathbb{E}\cro{\mathcal{T}_k\,|\,\mathcal{F}_{k-1}}}^2 + \norm{\frac{1}{n+1}\sum_{k=1}^n\Sigma_n -\mathbb{E}\cro{\mathcal{T}_k\,|\,\mathcal{F}_{k-1}}}^2 + \norm{\frac{1}{n+1}\Sigma_0}^2 \\
&= o\pa{\frac{(\ln n)^{1+\delta}}{n}} a.s.
\end{align*}

\bibliographystyle{apalike}
\bibliography{Biblio_Rapport}
\end{document}